\documentclass[letterpaper, 10 pt, conference]{ieeeconf}  

\IEEEoverridecommandlockouts                              

\usepackage{amsmath} 
\usepackage{amssymb}  
\usepackage{graphicx} 
\usepackage{booktabs}
\usepackage{multicol}
\usepackage{multirow}
\usepackage{courier}
\usepackage[table]{xcolor}  
\usepackage{array}  
\RequirePackage[colorlinks=true, allcolors=blue]{hyperref}
\usepackage[
    backend=biber,
    style=ieee,
    natbib=true,
    doi=false,
    url=false,
    isbn=false,
]{biblatex}

\AtEveryBibitem{%
   \clearfield{Title}%
   \clearfield{eprint}%
   \clearfield{note}%
}

\newbox{\myorcidaffilbox}
\sbox{\myorcidaffilbox}{\large\includegraphics[height=1.1ex]{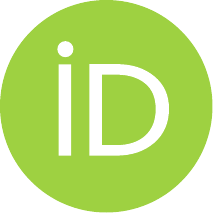}}
\newcommand{\orcidaffil}[1]{%
  \href{https://orcid.org/#1}{\usebox{\myorcidaffilbox}}}
  
\title{\LARGE \bf
Fusing uncalibrated IMUs and handheld smartphone video to reconstruct knee kinematics
}

\author{J.D. Peiffer$^{1,2,\orcidaffil{0000-0003-2382-8065}}$, Kunal Shah$^{1}$, Shawana Anarwala$^{1}$, Kayan Abdou$^{1}$  and R. James Cotton$^{1,3,\orcidaffil{0000-0001-5714-1400}}$
\thanks{This work was generously supported by the American Neurological Foundation, the Restore Center P2C (NIH P2CHD101913), and the Research Accelerator Program of the Shirley Ryan AbilityLab. JDP
is supported by the National Science Foundation Graduate Research Fellowship Program under Grant No. DGE-2234667.}
\thanks{$^{1}$ Center for Bionic Medicine, Shirley Ryan AbilityLab, 355 E Erie St, Chicago, IL
        {\tt\small jpeiffer@sralab.org}}%
\thanks{$^{2}$ Department of Biomedical Engineering, Northwestern University, Evanston, IL
}%
\thanks{$^{3}$ Department of Physical Medicine and Rehabilitation, Northwestern University, Evanston, IL
}%
}

\begin{document}

\maketitle
\thispagestyle{empty}
\pagestyle{empty}

\begin{abstract}

Video and wearable sensor data provide complementary information about human movement. Video provides a holistic understanding of the entire body in the world while wearable sensors provide high-resolution measurements of specific body segments. A robust method to fuse these modalities and obtain biomechanically accurate kinematics would have substantial utility for clinical assessment and monitoring. While multiple video-sensor fusion methods exist, most assume that a time-intensive, and often brittle, sensor-body calibration process has already been performed. In this work, we present a method to combine handheld smartphone video and uncalibrated wearable sensor data at their full temporal resolution. Our monocular, video-only, biomechanical reconstruction already performs well, with only several degrees of error at the knee during walking compared to markerless motion capture. Reconstructing from a fusion of video and wearable sensor data further reduces this error. We validate this in a mixture of people with no gait impairments, lower limb prosthesis users, and individuals with a history of stroke. We also show that sensor data allows tracking through periods of visual occlusion.

\end{abstract}

\section{INTRODUCTION}

The kinematics of human movement can provide important health information. For instance, upper limb kinematics change significantly in the months following a stroke \cite{thrane_upper_2020}, while lower limb kinematics can discriminate between healthy and pathological gait and changes during rehabilitation \cite{cotton_improved_2023}. Traditionally, measuring kinematics was performed in motion analysis laboratories using marker-based methods \cite{doi:https://doi.org/10.1002/9780470549148.ch3}. This technique detects reflective markers placed on the surface of the skin, which are then fit with a biomechanical model to obtain joint angle kinematic trajectories. While this method produces accurate tracking, the marker placement requires significant domain knowledge and setup time that limits the usability of this technique in the clinic. 

\begin{figure*}
    \centering
    \includegraphics[width=0.95\linewidth]{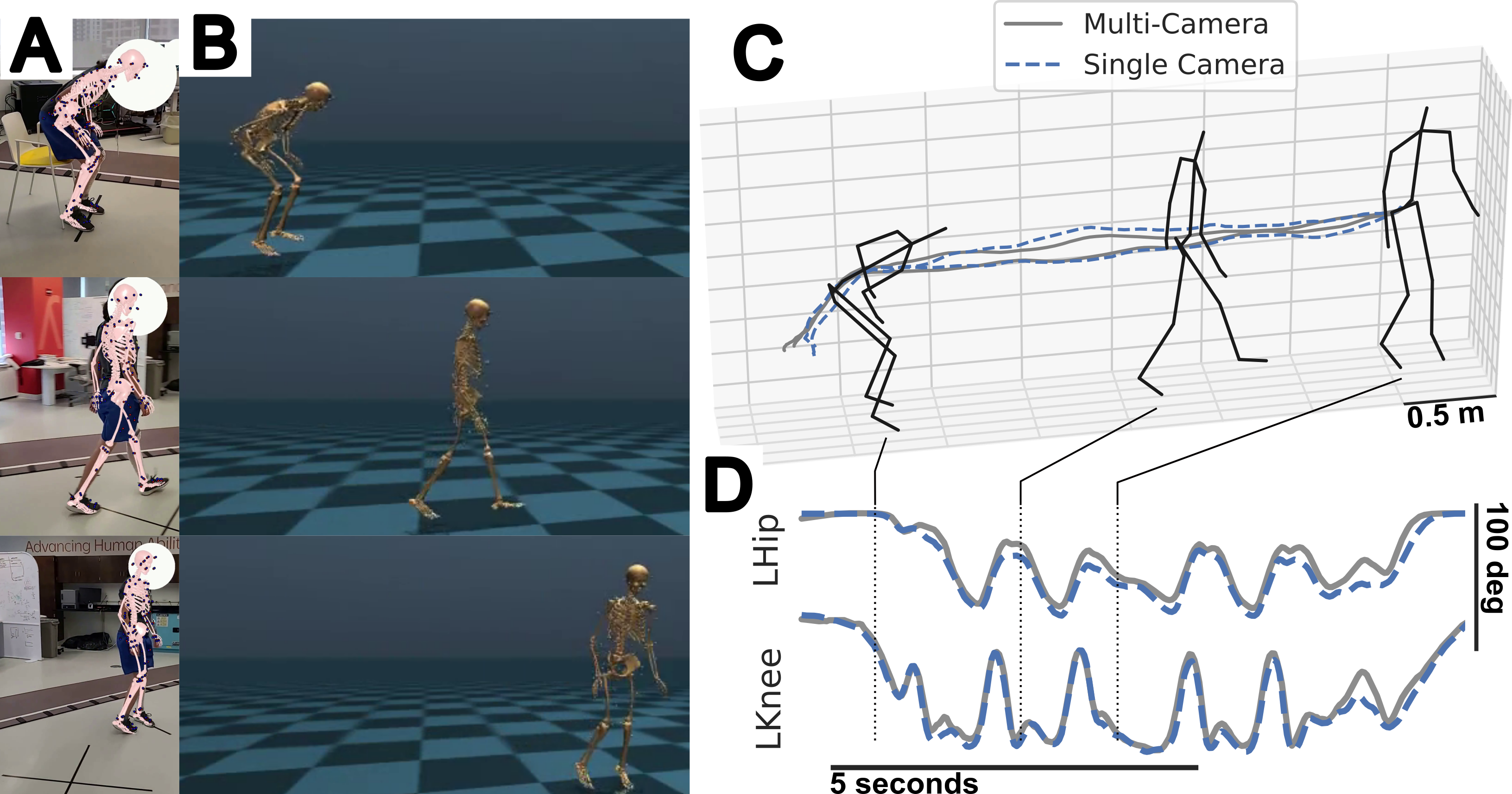}
    \caption{Using video from a handheld moving camera (A), our method disentangles camera and body orientation changes to accurately track a timed up and go test in global space (B,C). We validated the joint kinematics against a multi-camera system (D), finding close agreement between the two methods.
    }
    \label{fig:example}
    \vspace{-1.75em}
\end{figure*}
In recent years, multiview markerless motion capture has seen considerable improvement and expansion. This technology leverages computer vision to estimate joint locations from 2D RGB videos captured by multiple calibrated and synchronized cameras. Instead of physical surface markers, virtual marker locations, known as 'keypoints', are detected. As with marker-based systems, a biomechanical model can be fit to follow these virtual markers. Recent work has shown markerless systems achieve comparable accuracy to marker-based accuracy \cite{kanko_inter-session_2020} and our previous work has demonstrated its applicability to gait analysis in clinical populations \cite{cotton_optimizing_2023}. We have also shown that using differentiable biomechanical models allows us to formulate the biomechanical reconstruction from videos as an end-to-end optimization problem. Directly optimizing kinematic trajectories to fit the detected keypoints outperforms the traditional two-stage approach, which first computes marker trajectories and then applies inverse kinematics \cite{cotton_differentiable_2024}.

Historically, human pose estimation (HPE) has focused on detecting 2D keypoints in the image plane. However, contemporary methods such as MeTRAbs-ACAE now enable precise 3D keypoint estimates directly from images \cite{sarandi_learning_2022}. Concurrently, another approach involves using established mesh body models, specifically the Skinned Multi-Person Linear (SMPL) model, to optimize or directly regress shape and pose parameters \cite{loper_smpl_2015},\cite{kanazawa_end--end_2018}. 
The shaping and forward kinematic processes of this model are differentiable and benefit from GPU acceleration, rendering them suitable for optimization or regression techniques. However, the kinematic tree configuration does not accurately represent anatomical joint locations and articulations \cite{keller_skin_2023}, thus restricting the model's applicability in biomechanics where specific conventions are established \cite{wu_isb_2002}.
In this work, we extend our previous work optimizing a differentiable biomechanical model \cite{cotton_differentiable_2024} to 3D keypoints detected from monocular videos. We also show how this framework can be extended to fuse video data with wearable sensor data, even if the sensor placement has not been calibrated.

Wearable sensors, particularly Inertial Measurement Units (IMUs) have been used with vision-based approaches to enhance human pose estimate accuracy \cite{desmarais_review_2021,li_visualinertial_2023}. IMUs include an accelerometer, gyroscope, and magnetometer to measure linear acceleration, angular velocity, and heading relative to magnetic north, respectively.  These measurements are fused using algorithms like the Kalman or complementary filter to estimate orientation \cite{cotton_wearable_2019},\cite{mahony_nonlinear_2008}. Sensor data can be collected at much higher rates than video, and when fused with stationary video, they have shown significant improvements in pose estimation \cite{shin_markerless_2023}. The approaches are complementary: video tracks the global configuration of a person, while IMUs provide high-resolution information that is robust to occlusions. However, extracting biomechanics from sensor data typically requires a cumbersome calibration process, making it impractical for routine clinical use.

There have been a few recent approaches to creating portable motion capture and biomechanics systems. First, OpenCap \cite{uhlrich_opencap_2023} uses two or more calibrated, stationary smartphones on tripods to capture RGB images and provides a convenient web application for users to process their data. OpenSense \cite{al_borno_opensense_2022} uses 8 IMUs to calculate the kinematics and kinetics of a biomechanical model. Setup and calibration steps, required for both of these, limit the adoption of motion capture and biomechanics into routine clinical use \cite{mitchell_factors_2023}. We have previously described our portable smartphone system for capturing synchronous sensor and video data during gait in the clinic \cite{cimorelli_validation_2024}. In that work, sensor data was only used to detect gait cycles to validate the vision algorithms. Here we leverage the data from our Portable Biomechanics Laboratory to fuse data from a single mobile camera and uncalibrated wearable sensors.

Our work addresses two critical gaps: (1) the absence of methods to estimate biomechanically-grounded kinematics using a single handheld camera, and (2) the lack of techniques for utilizing uncalibrated IMU data in kinematic tracking, which is essential for their use in time-constrained clinical settings.
The primary contributions of this work are:

\begin{itemize}
    \item We extend our differentiable biomechanics approach \cite{cotton_differentiable_2024} from multiview markerless motion capture data to monocular video, which already produces accurate fits.
    \item We show that uncalibrated IMU data can be fused with smartphone video, improving the accuracy of our knee kinematics.
    \item We demonstrate global orientation tracking of the participant, disentangling the camera and participant rotation, using both rotations measured from the sensor and smartphone gyroscope (Figure \ref{fig:example}).
\end{itemize}

\section{METHODS}

\subsection{Data collection}
This study was approved by the Northwestern University Institutional Review Board. 38 participants participated in this study, of those 38, 7 participants were lower limb prosthesis users, 11 had a history of stroke and 20 were able-bodied controls. All recordings took place in 7.4 x 8 m room with participants walking diagonally across the room (11 m) for one trial. Participants completed nine walking trials, 3 at a self selected pace, 3 at a fast, and 3 at a slow pace. Gait data was collected simultaneously with both our wearable sensor and smartphone platform as well as our multiview markerless motion capture (MMMC) system. We analyzed 1-2 videos of each participant.

\subsection{Portable Biomechanics Laboratory}

Our Portable Biomechanics Laboratory (PBL) system consists of a Samsung Galaxy S20 smartphone and custom-made wearable IMUs \cite{cimorelli_portable_2022,cotton_wearable_2019}. The smartphone collects 1080 $\times$ 1920 resolution RGB video data at 30 Hz and internal smartphone gyroscope data at 100 Hz. It also records depth videos, which were not used in this study. The IMUs collect accelerometer, magnetometer, and gyroscope data at 562.5 Hz. They run a complementary filter that estimates orientation data at 55 Hz \cite{cotton_wearable_2019}.  The gyroscope and orientation data are used in our fusion. IMUs were placed unilaterally on the participant’s thigh and lower leg using velcro straps, typically on the lateral side. For participants with unilateral gait impairments, sensors were placed on the impaired limb. For each trial, the researcher followed behind the participant holding the smartphone as the participant walked across the room (Fig \ref{fig:example}).

\subsection{Multiview Markerless Motion Capture}

The Multiview Markerless Motion Capture (MMMC) system consists of 8-12 FLIR BlackFly S GigE cameras which acquire synchronized RGB video at 29 fps. Cameras were arranged such that at least three cameras covered participants at all points during a recording. A detailed description of this system can be found in \cite{cotton_markerless_2023-2}. We reconstructed biomechanical fits using methods described in \cite{cotton_differentiable_2024}. This uses the same biomechanical model used in the monocular reconstruction to provide a consistent basis for comparison.

\subsection{Video processing}
We processed smartphone and MMMC videos using PosePipe \cite{cotton_posepipe_2022}, an open-source tool for human pose estimation. Bounding boxes were first created using DeepSort \cite{wojke_simple_2017} and 3D keypoints were estimated using MeTRAbs-ACAE \cite{sarandi_learning_2022}. This algorithm produces the superset of many keypoint sets, and we used the 87 BML-MoVi keypoints \cite{ghorbani_movi_2021}, as this dense markerset is designed for biomechanics. As recommended by the MeTRAbs-ACAE author, we computed keypoint confidence by measuring the standard deviation of each joint location estimated from 10 different augmented versions of each video frame. This was converted to a confidence estimate using a sigmoid function with a half maximum at 30mm and a width of 10mm. We annotated the participant of interest if multiple participants were present in the video. 

\subsection{Notation}
We follow the convention presented in \cite{lynch_modern_2017} for rotations. In brief, rotation $R_{ab}$ represents a rotation from reference frame $\{a\}$ to reference frame $\{b\}$. If this rotation varies with time, as in the case of a body segment or sensor reading, we represent that rotation as $R_{ab}(t)$. We denote arbitrary vector quantities that exhibit temporal variation as $\vec{V}(t)$, where $\vec{V}$ represents the vector in question and $t$ indicates its dependence on time. We represent quantities that are estimates of our model with the $\hat{ \cdot}$ symbol and measured quantities, such as video-estimated 3D keypoints and raw sensor readings, with no hat.

\subsection{Body model}

For both the multi and single camera data, we used a biomechanically grounded, forward kinematic model implemented in MuJoCo \cite{todorov_mujoco_2012},\cite{caggiano_myosuite_2022}. A full summary of this model can be found in \cite{cotton_differentiable_2024}. In brief, this model is based on a MuJoCo implementation \cite{al-hafez_locomujoco_2023} of an OpenSim model \cite{hamner_muscle_2010}. We further modified it to include a neck joint and optimized the default marker locations for the BML-MoVi keypoints \cite{ghorbani_movi_2021}. Forward kinematics of this model can be massively accelerated and parallelized on the GPU with MuJoCo MJX. Scaling within the model is executed via a matrix that maps a set of scale parameters to the isotropic scaling of individual body segments. We defined eight scale parameters: overall size, the pelvis, left thigh, left leg and foot, right thigh, right leg and foot, the left arm, and the left leg. We also included an $87 \times 3$ matrix of marker offsets for small changes in the 87 MoVi keypoint locations. Together, these parameters are represented as a vector $\vec \beta \in \mathbb{R} ^{8+87 \times 3}$. The model inputs, denoted by $\vec \theta \in \mathbb{R}^{40}$, include six parameters for translating and orienting the pelvis, and 36 parameters for actuating the remaining segments of the kinematic tree. We assigned each of the 87 MoVi keypoints to specific body segments and positioned them using the forward kinematic process, as defined by the following equation:

\begin{equation}
\hat p_n(t) = \mathcal M_p(\vec \beta,\vec \theta(t))
\label{eq:FK1}
\end{equation}

In this equation, $\vec \beta$ and $\vec \theta$ represent the scale and pose parameters, respectively. The term $\hat p_n \in \mathbb{R}^{87 \times 3}$ denotes the 3D locations of the 87 MoVi keypoints relative to the body model, which is oriented within a global frame ${n}$.

Additionally, the forward kinematic process calculates the orientations of each body segment $b_i$ in the global coordinate system:

\begin{equation}
\hat R_{nb}(t)=\mathcal M_{R}(\vec \beta,\vec \theta(t))
\label{eq:FK2}
\end{equation}

In this model, the orientation of an arbitrary body segment $R_{nb_i} \in \mathbb{R}^{3 \times 3}$ is represented within a matrix $R_{nb} \in \mathbb{R}^{22 \times 3 \times 3}$, encapsulating the orientations of all segments.

\subsection{Fusion Architecture}

\begin{figure*}
    \centering
    \vspace{1em}
    \includegraphics[width=\linewidth]{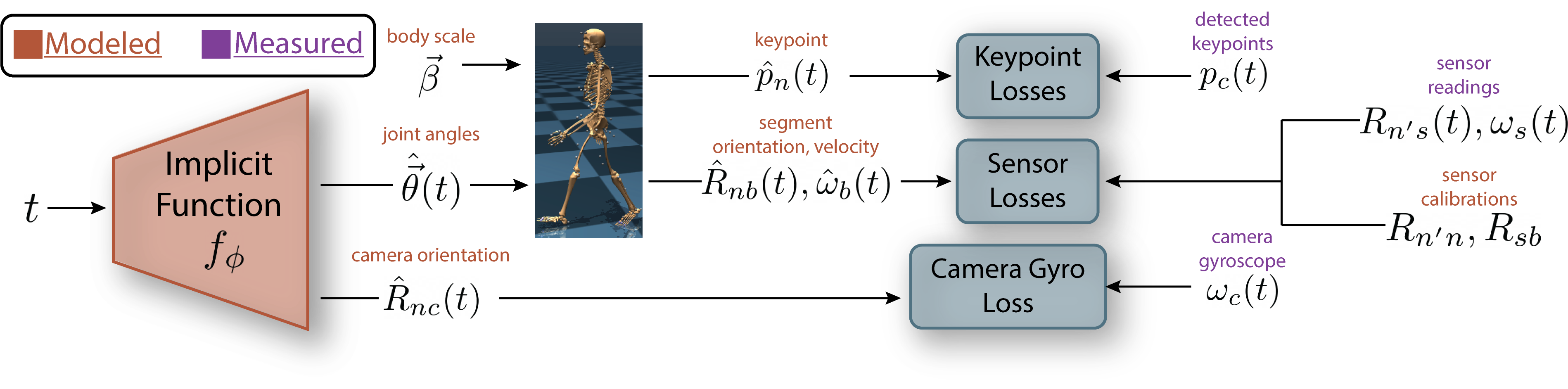}
    \vspace{-3.5em}
    \caption{Our method jointly optimizes an implicit function learning the trajectory of an individual recording and $\vec \beta$ the scaling parameters of a biomechanical model. The implicit function takes time as an input and outputs $\vec \theta(t)$, the joint pose parameters, and $\hat R_{nc}(t)$ the orientation of the camera in a global frame. The pose and scaling parameters produce joint locations $\hat p_n(t)$ and joint orientations $\hat R_{nb}(t)$ that are compared to detected keypoints and sensor readings. Note the sensor calibrations are included in the optimziation. Changing the keypoint reference frame using $\hat R_{nc}$, as in Eqs. \ref{eq:keypoint_loss} and \ref{eq:reprojection_loss}, is not explicitly shown.
    }
    \label{fig:layout}
    \vspace{-1.5em}
\end{figure*}

A diagram of this architecture is shown in Figure \ref{fig:layout}. Based on recent work employing an implicit function triangulating 2D markers and biomechanical poses \cite{cotton_optimizing_2023,cotton_differentiable_2024}, we use an implicit function $f_\phi$ to optimize a mapping from recording time $t$ to poses $\vec \theta(t)$, implemented as a multi-layer perceptron:

\begin{equation}
    f_\phi:t \rightarrow (\hat{\vec {\theta}} 
    ,\hat R_{nc})
    \label{eq:net}
\end{equation}

The pose vector $\hat{\vec{\theta}}$ is passed through the forward kinematic model \ref{eq:FK1},\ref{eq:FK2}. The implicit function also outputs $\hat{R}_{nc}(t)$ representing the smartphone camera’s orientation relative to the global north frame $\{n\}$ at time $t$.

Composing equations \ref{eq:FK1} or \ref{eq:FK2} with equation \ref{eq:net} yields a kinematic trajectory, parameterized by $\phi$, that outputs estimated 3D joint locations, body segment orientations, and camera orientations at an arbitrary timepoint in the recording trial.

To obtain the rate of change of model outputs, such as predicted gyroscope measurements, we differentiate through the forward process represented in equation \ref{fig:layout} to obtain $\hat{\dot{R}}_{nb_i}$ and the forward pass in equation \ref{eq:net} for an estimate of $\hat{\dot{R}}_{nc}$.

\subsection{Sensor Calibration Solution}

IMUs produce an orientation estimate in their own reference frame $\{n'\}$. Here, we are defining $\{n\}$ as a horizontal reference frame facing north, and $\{n'\}$ captures any slight heading offset from the IMU zero position due to magnetic disturbance or imperfect sensor magnetic calibration. We denote the measured IMU orientation of sensor $i$ at a point in time as $R_{n's_i}(t)$ where $\{s_i\}$ is the sensor frame for a sensor attached to segment ${i}$. We transform this reading onto its body segment $b_i$ in the global frame with equation \ref{eq:sensor_cal}. If our calibration and model are optimized correctly, the calibrated sensor and body model should agree (equation \ref{eq:sensor_approx}).
\begin{gather}
    R_{nn'}(t)R_{n's_i}(t)R_{sb_i} \approx \hat R_{nb_i}(t)    \label{eq:sensor_cal} \\
    R_{nn'}(t)R_{n's}(t)R_{sb_i} \approx \mathcal M_{R_i}(\hat{\vec{\beta}},\hat{\vec{\theta(t)}})\label{eq:sensor_approx}
\end{gather}
\noindent with $R_{sb_i}$  being the rotation from the sensor frame to the body segment frame and $R_{nn'}$ a rotation from the global frame to the IMU's home frame. Critical to our approach, we include $R_{nn'}$ and $R_{sb_i}$ as learned parameters in our optimization. 

While the orientation estimation algorithm is designed to minimize drift \cite{cotton_wearable_2019}, the frame $\{n'\}$ may still drift over time. We account for this by representing $R_{nn'}(t)$ as a piecewise spherical linear interpolation (SLERP) \cite{shoemake_animating_1985} between three optimized quaternions $q_{nn'}(0)$, $q_{nn'}(T_{mid})$ and $q_{nn'}(T)$. 


 
The sensor gyroscope reading $\omega_{s_i}$ for a sensor on segment $i$ depends on the body segment alignment. We differentiate through our implicit function and \ref{eq:FK2} with respect to time using autograd to obtain $\hat{\dot{R}}_{nb_i}$. Using the sensor calibration, $R_{sb_i}$, we predict sensor $i$'s gyroscope reading as:

\begin{equation}
    R_{sb_i}( \hat{R}_{nb_i}^{-1} \hat{\dot{R}}_{nb_i})=[\hat{\omega}_{s_i}]
    \label{eq:gyro_sens}
\end{equation}
\noindent where $[\hat{\omega}_s]$ is the skew-symmetric representation of the gyroscope reading. For a derivation of this equation refer to \cite{lynch_modern_2017}.
Similarly, we predict the phone gyroscope with:
\begin{equation}
    (\hat R_{nc}^{-1}\hat{\dot{R}}_{nc})=[\hat{\omega}_{c}]
    \label{eq:gyro_cam}.
\end{equation}

\subsection{Loss functions and implementation}
A forward pass of our model produces estimates we denote as $\hat p_n(t)$, $\hat R_{nb}(t)$, $\hat R_{nc}(t)$, $\hat \omega_s(t)$, and $\hat \omega_c(t)$. We optimize the parameters of the neural network $\phi$, the sensor calibrations $R_{sb}$, $R_{nn'}$ and the body scaling parameters $\vec \beta$ using the following losses. To reduce notational clutter, we define our losses at a single point in time, but all of these terms are averaged over time except for the sensor calibrations.

\subsubsection{Keypoint loss}
We obtain a pure video estimate of the 3D keypoint locations  $p_c$ using MeTRAbs-ACAE \cite{sarandi_learning_2022}. These keypoints are in the camera reference frame. We can rotate them into the global frame using $p_n=\hat R_{nc}p_c$ and define a loss function on the euclidean distance between keypoints from the model $\hat p_n$ with the video keypoints as

\begin{equation}
    \mathcal L_{keypoint}(\phi,\vec \beta)=\frac{1}{J}\sum_{j\in J} c(j) g(\Vert \hat p_n - \hat R_{nc}p_{c} \Vert_2)
    \label{eq:keypoint_loss}
\end{equation}

where $c(j) \in [0,1]$ is the confidence score for joint keypoint $j$ which also varies with time and $g(\cdot)$ is a Huber loss which is quadratic within 100 cm and linear after, which was necessary for stabilizing early training. This loss is computed between the 3D keypoints set with their mean translation removed.

\subsubsection{Reprojection loss}
 MeTRAbs-ACAE also produces 2D keypoints in the image frame $\vec x \in \mathbb{R}^{87 \times 2}$. The 3D keypoints produced by our model $\hat p_n$ can be rotated into the camera frame using $\hat R_{nc}$ and projected through a camera function $\Pi$ with the calibrated camera intrinsics, to compute the error with detected 2D keypoints:

 \begin{equation}
    \mathcal L_{reprojection}(\phi,\vec \beta)=\frac{1}{J}\sum_{j\in J} c(j)g( \Vert \Pi(\hat R_{nc}^{-1}\hat p_n) - \vec x \Vert_2)
    \label{eq:reprojection_loss}
\end{equation}
In this loss, $g(\cdot)$ is a Huber loss which is quadratic within 100 pixels and linear after.
 
\subsubsection{Sensor attitude loss}

The forward kinematic process produces an estimate for each body segment $i$'s orientation in the world frame $\hat R_{nb_i}$. For each segment with a sensor attached $B_s$, We transform the sensor reading $R_{n's_i}$ to this frame using equation \ref{eq:sensor_cal} and measure the angular difference between the two estimates with the following loss:

\begin{equation}
\begin{split}
\mathcal L_{att}(\phi,\vec \beta,R_{sb_i},R_{n_i'n}) & =\\ 
& \frac{1}{|B_s|}\sum_{i \in B_s} (\frac{\text{arccos Tr}[\hat{R}_{nb_i}R_{nb_i}^{-1}]-1}{2})^2
\label{eq:orientation_loss}
\end{split}
\end{equation}

\noindent, where $B_s$ represents the set of body segments with sensors attached to them.

\subsubsection{Sensor gyroscope loss}

We compare the estimate of the sensor gyroscope obtained from our model $\hat{\omega}_{s_i}$ (equation \ref{eq:gyro_sens}) to the measured sensor gyroscope measurement $\omega_{s_i}$ for each sensor $s_i$ using the mean squared error:

\begin{equation}
    \mathcal L_{gyro\_sens}(\phi,\vec \beta,R_{sb_i})=\frac{1}{|B_s|}\sum_{i\in B_s} \Vert\hat \omega_{s_i} - \omega_{s_i}\Vert^2_2
\label{eq:gyro_loss}
\end{equation}

\subsubsection{Camera gyroscope loss}
Finally, we include a loss between the predicted camera gyroscope reading $\hat \omega_c$  from equation \ref{eq:gyro_cam} and the measured camera gyroscope reading $\omega_c$ using the mean squared error:

\begin{equation}
    \mathcal L_{gyro\_phone}(\phi)= \Vert\hat \omega_{c} - \omega_{c}\Vert^2_2
    \label{eq:l_gyro_cam}
\end{equation}
\subsection{Implementation}
 Implementation is done using Jax \cite{bradbury_jax_2024}, Equinox \cite{kidger_equinox_2021} and Optax. We combine three different optimizers using Optax's  \texttt{multi\_transform}. First, we use an Adam optimizer with weight decay set to $1e-5$, exponential learning rate decay starting at $1e-4$ and ending at $1e-5$, and $\beta=(0.9,0.8)$ for the implicit function.  After 10,000 iteration steps, we begin optimizing the sensor calibrations with an Adam optimizer with a learning rate $1e-5$. Finally, we optimize the time offsets between the IMU and video data with a third Adam optimizer with a learning rate $1e-4$, $\beta_1=0.85$. We found that annealing the weights of the sensor losses was crucial for stability. An optimization for a $\sim$10 second video converges after 20000 optimization steps and runs at 50 iterations/s on an NVIDIA A6000 GPU. Since our implicit function $f_\phi(t)$ (equation \ref{eq:net}) can be evaluated at arbitrary timepoints, we were able to train our model using the full resolution sampling rate of all modalities. A training step computes losses over 500 random samples of each data source.

 \subsection{Evaluation}
 To evaluate the agreement between the camera and sensor system, we calculated the residuals between our fused model's prediction and measured values averaged over the entire recording. This included:

 \begin{enumerate}
     \item \textit{Keypoint Residual}: Mean Euclidean distance between the 3D keypoints detected from MeTRAbs-ACAE $p_c$ and 3D keypoints predicted by our model and rotated into the camera frame $R_{nc}^{-1}\hat p_n$.
     \item \textit{Reprojection Residual}: Mean distance in pixels between the 2D keypoints detected from MeTRAbs-ACAE $x$ and the model predictions rotated into the camera frame and projected through the smartphone camera model $\Pi(R_{nc}^{-1}\hat p_n$).
     \item  \textit{Phone gyroscope Residual}: Mean difference between the phone gyroscope predicted by our model $\hat \omega_c$ and gyroscope measured by the smartphone $\omega_c$. Calculated using \ref{eq:l_gyro_cam}.
     \item \textit{Sensor Gyroscope Residual}: Mean difference between the sensor gyroscope predicted by our model and raw gyroscope readings from the sensors, averaged across all sensors. Calculated using \ref{eq:gyro_loss}.
     \item \textit{Attitude Residual}: Mean difference between attitude predicted by our model $\hat R_{nb_i}$ and raw sensor attitude readings $R_{nb_i}$ averaged across all sensors.  Calculated using \ref{eq:orientation_loss}.
 \end{enumerate}
 
Next, we synchronized each PBL recording to the MMMC system. As the MMMC and PBL systems were collected at different sampling rates, we evaluated our model at the MMMC timepoints for comparison. As sensors were placed above and below the knee, the knee angle was our primary measure of accuracy between our fusion solution and multi-camera system. The metrics comparing the fusion and multi-camera system are:
 \begin{enumerate}
     \item \textit{Pearson’s correlation coefficient} measures the covariation of the flexion angles calculated from the fusion solution and MMMC system.
     \item \textit{Mean absolute error (MAE)} measures the absolute angular error of the flexion angle between the fusion solution and MMMC system.
     \item \textit{Mean absolute error mean adjusted (MAE-MA)} removes the mean value of each kinematic trace before comparing the comparing the absolute angular error. 
 \end{enumerate}
Summary statistics are presented as: \texttt{Median (IQR)}. To test for significant differences in these metrics, we used a Wilcoxon signed-rank test, defining significance as $p<0.05$. 

\emph{Occlusion Experiments.} To estimate how well our method tracked kinematics from the sensors when video is partially occluded, we artificially occluded the sensorized leg for the middle half of each recording. We allowed our optimization to proceed as above and compared the same metrics. 

\begin{figure}
    \centering
    \includegraphics[width=\linewidth]{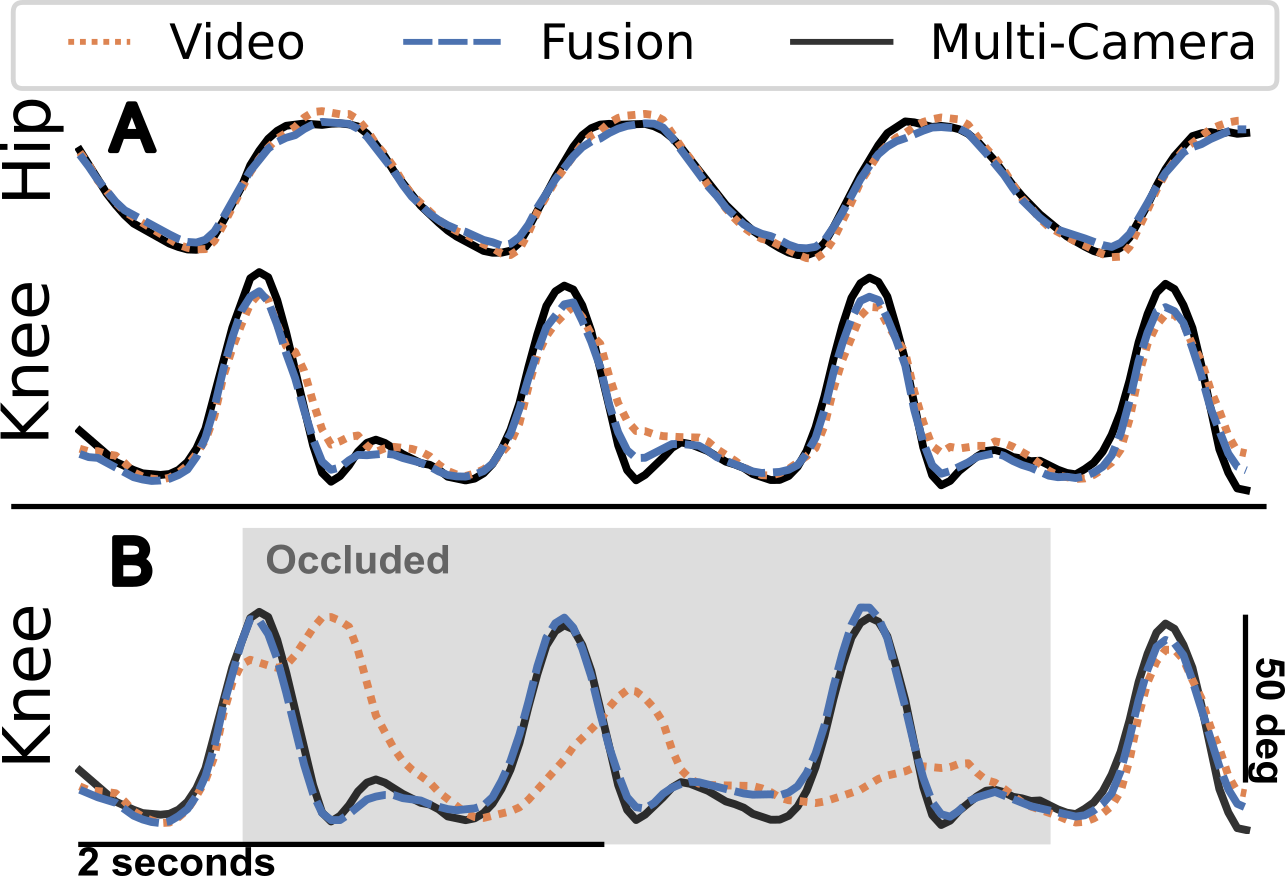}
    \caption{Example lower limb kinematics during walking. (A) With no occlusions, both video and fusion track closely with multi-camera estimates. In this example, the fusion solution tracks the knee better during heel strike. (B) The fusion solution continues to track knee kinematics during occlusion while the video does not.}
    \label{fig:knee_angles}
    \vspace{-1.75em}
\end{figure}

\section{RESULTS}

\subsection{Qualitative Results}
Since this system does not require an explicit calibration trial or static pose, individuals with gait or balance impairments were able to complete this experiment with relative ease. A recording session consisting of 9+ walking trials was typically completed in $<$15 minutes, with a majority of the setup time attributed to EMG placement (data not used in this study). In some trials, a physical therapist was closely following a participant or even holding a gait belt, however, the person and keypoint detection done using \cite{cotton_posepipe_2022} and \cite{sarandi_learning_2022} was generally robust to this. While the loss functions were able to track sensor readings quite closely, the relative weighting/annealing schedules of the individual loss terms proved to be very important in stable convergence and accurate results.

\begin{table*}
\caption{\scriptsize Fit residuals (Median (IQR)) across population and fit type}
\label{tab:sensor_alignment}
\begin{center}

\begin{tabular}{ll|lllll}
\toprule
& & \multicolumn{5}{c}{Residual}  \\
Population & Fit &  Keypoint (cm) & Reprojection (pixels) & Phone Gyroscope (deg/s) & Gyroscope (deg/s) & Attitude (deg) \\
\midrule
\multirow[t]{2}{*}{Control} & Video & 0.7 (0.08) & 5.35 (1.06) & 1.76 (1.54) & - & - \\
& Fusion & 0.76 (0.08) & 5.52 (0.97) & 1.73 (1.51) & 23.59 (6.17) & 3.68 (1.01) \\
\cline{1-7}
\multirow[t]{2}{*}{Stroke} & Video & 0.68 (0.09) & 4.72 (0.72) & 2.18 (1.09) & - & - \\
& Fusion & 0.75 (0.07) & 4.93 (0.72) & 2.15 (1.11) & 21.8 (13.6) & 4.68 (1.36) \\
\cline{1-7}
\multirow[t]{2}{*}{Prosthetic} & Video & 0.69 (0.08) & 4.74 (0.57) & 2.34 (0.54) & - & - \\
& Fusion & 0.78 (0.13) & 4.97 (0.64) & 2.3 (0.54) & 23.03 (6.38) & 4.44 (1.25) \\
\cline{1-7}
\bottomrule
\end{tabular}
\end{center}
\end{table*}

\subsection{Fusion Consistency}
Following fusion, residuals errors were quite low (Table \ref{tab:sensor_alignment}). Our model was capable of closely following the detected keypoints with an average of $< 1$ cm 3D keypoint error across all groups and $\sim 5$ pixels of average reprojection error.  The keypoint residuals were generally lowest in the video-only fit and greater when the wearable sensors were added to the optimization. This is not unexpected as the sensors may compensate for incorrectly detected keypoints. The phone gyroscope estimate tracked within $\sim 2$ deg/s for all conditions. Figure \ref{fig:example} shows an example of the model disentangling the phone rotation, supervised with the measured phone gyroscope, and the pelvis rotation, which, in this case, accurately recovers the global translation within 6 cm. Sensor gyroscope residuals averaged 21-23 degrees per second. We note that while this number is quite high, the relative weighting of the gyroscope error in the optimization was lower than that of other terms. Attitude residuals were lowest in the control population $\sim 3$ degrees and higher in the prosthetic and stroke groups.

\subsection{Multi-Camera Comparison}
Across all 60 videos recorded of the 38 subjects, video estimates produced a MAE-MA of 2.42 (1.2) degrees and 3.91 (1.55) degrees for hip and knee, respectively. These estimates were significantly decreased to 2.16 (0.86) degrees and 2.9 (1.27) degrees for the hip and knee, respectively. Pearson's correlation coefficients were  high for video, 0.98 (0.02) for hip and 0.97 (0.06) for knee, with significant increases with the addition of sensors: 0.99 (0.01) for hip and 0.98 (0.02) for knee. This indicates a high temporal consistency between our estimates and ground truth. A summary of these results can be found in Table \ref{tab:mmc_compare} and example traces in Figure \ref{fig:knee_angles}. Sensors significantly decreased knee MAE-MA in all groups. Control populations had the lowest MAE, MAE-MA and highest correlations.

\subsection{Artificial Video Occlusion}
\begin{figure}
    \centering
    \includegraphics[width=\linewidth]{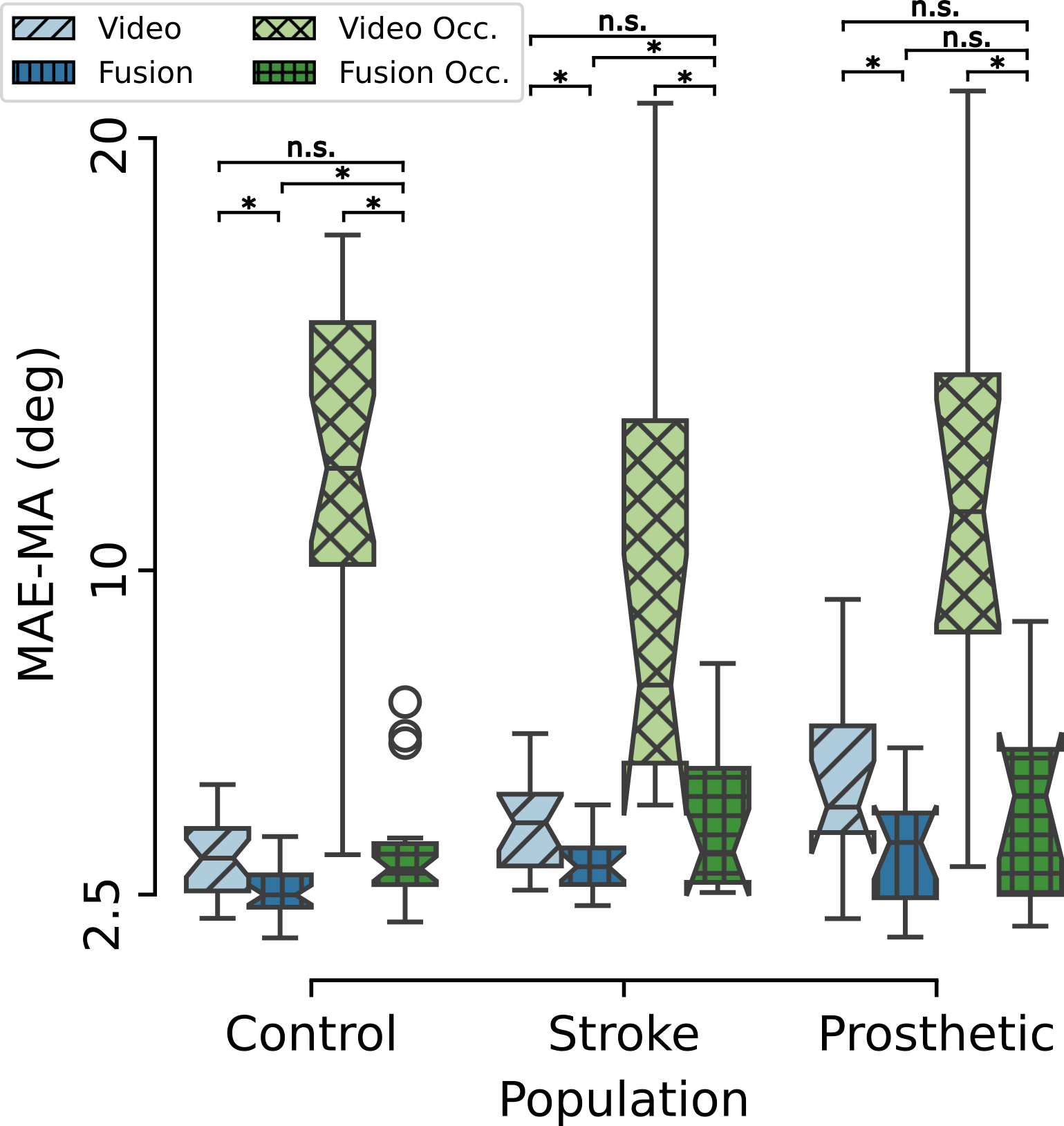}
    \caption{Distribution of knee mean adjusted MAE (MAE-MA). Addition of sensors generally decreases MAE. Artificial occlusion of the sensorized leg increased MAE in video-only fits, yet fusion fits significantly decreased the error due to this. Fusion fits during occlusion were not significantly different from video fits.  
    }
    \label{fig:distributions}
    \vspace{-2em}
\end{figure}

Our results, depicted in Figures \ref{fig:knee_angles} and \ref{fig:distributions}, demonstrate the impact of occlusion on MAE-MA. Specifically, occlusions increased MAE in the video-only approach, while the integration of sensors mitigated this effect. Fusion fits during occlusion were not significantly different from video fits. Additionally, Pearson correlations significantly dropped in the video-only scenario during occlusion (0.51 (0.41)), but were restored with the use of sensors (0.97 (0.05)).

\begin{table*}
\caption{\scriptsize Comparison of fusion solutions with ground truth (Median (IQR)). Bold for significance vs video.}
\label{tab:mmc_compare}
\begin{center}
\begin{tabular}{ll|ll|ll|ll}
\toprule
& & \multicolumn{2}{c}{$\downarrow$MAE (deg)} & \multicolumn{2}{c}{$\downarrow$MAE-MA (deg)} & \multicolumn{2}{c}{$\uparrow$Pearson's R}  \\
Population & Fit & Hip & Knee & Hip & Knee & Hip & Knee \\
\midrule
\multirow[t]{2}{*}{Control} & Video & 3.49 (3.28) & 4.3 (2.51) & 1.99 (0.7) & 3.35 (1.48) & 0.99 (0.01) & 0.98 (0.02) \\
& Fusion & 4.02 (3.31) & \textbf{3.44 (1.73)} & 1.94 (0.66) & \textbf{2.53 (0.76)} & 0.99 (0.0) & \textbf{0.99 (0.01)} \\

\cline{1-8}
\multirow[t]{2}{*}{Stroke} & Video & 3.96 (2.35) & 4.83 (2.28) & 2.52 (1.32) & 4.17 (1.57) & 0.98 (0.03) & 0.94 (0.15) \\
& Fusion & 3.96 (1.52) & \textbf{4.48 (2.48)} & 2.4 (0.96) & \textbf{3.06 (0.91)} & 0.98 (0.02) & \textbf{0.97 (0.08)} \\
\cline{1-8}
\multirow[t]{2}{*}{Prosthetic} & Video & 4.77 (3.04) & 6.08 (3.3) & 3.19 (1.32) & 4.48 (3.4) & 0.97 (0.01) & 0.94 (0.05) \\
& Fusion & \textbf{4.64 (2.51)} & 5.71 (4.2) & \textbf{2.77 (1.52)} & \textbf{3.65 (1.92)} & \textbf{0.98 (0.03)} & \textbf{0.97 (0.05)} \\
\cline{1-8}
\bottomrule
\end{tabular}
\end{center}
\end{table*}

\section{DISCUSSION}
While computer vision and wearable sensors offer much promise for rehabilitation and clinical monitoring, they remain primarily a research tool that has not translated to standard clinic or home monitoring. Ease of setup, use, and troubleshooting is fundamental to the adoption of technologies in the clinic. Clinicians do not have time to utilize a technology unless it is immediately intuitive and “foolproof” \cite{mitchell_factors_2023}. The same holds for patients using technology at home for therapy \cite{chen_home-based_2019}.  The objective of this work was to develop a method for accurate kinematic tracking from a handheld camera/sensor system with no calibration requirements. We did this by jointly optimizing an implicit function that fits the pose and camera trajectory as well as the sensor-body calibration rotations. This approach provides accurate measurements of lower body kinematics across control, lower limb prosthesis user, and neurological populations. The accuracy of the knee is within 6 degrees for video alone when comparing to MMMC, with errors less than 4 degrees when removing the bias, which is already quite good. The inclusion of wearable sensors reduces the measurement error by another degree. Our dataset contained a large amount of walking data, so we chose to focus on knee kinematics in this study, however, this method should generalize to other joints as well. 

The results in Table \ref{tab:sensor_alignment} demonstrate that our algorithm is able to solve for a solution to Equation \ref{eq:sensor_cal} that rotates raw IMU data into the body frame. We found that the phone gyroscope was useful for disentangling the camera vs root rotation (Figure \ref{fig:example}). We were surprised at how high the sensor gyroscope residuals were, particularly as preliminary experiments with SMPL models showed lower gyro residuals. We experimented with different weighting/annealing schedules of the gyroscope loss; however, our model did not allow for closer fitting without disturbing the kinematic trajectory. An explanation for this could be that our model did not include varus/valugs rotation at the knee joint, which could be incorporated in the future. Alternatively, numerical errors from differentiating through the forward kinematics could have limited the accuracy (Equations \ref{eq:gyro_sens},\ref{eq:gyro_cam}). All other residuals we found to be quite low, $< 1$ cm and 5 pixels for keypoints, $\sim 2$ deg/s of camera gyroscope error, and  $\sim 4$ degrees of attitude error. 

Table \ref{tab:mmc_compare} shows that our fusion method produces accurate kinematics for our joint of interest when compared to a multi-camera system \cite{cotton_markerless_2023-2,cotton_differentiable_2024}. We were pleased with how small the joint angle error were with a video-only approach with 3.35, 4.17, and 4.48 degrees of knee MAE-MA for control, stroke and prosthetic populations, respectively. This indicates that for many applications, video alone could be sufficient. Despite the high bar set by video, the addition of the wearable sensors improved kinematic tracking for all populations by 0.8 - 1.1 degrees. 
Our synthetic occlusion experiments also show an additional benefit of sensors, where the fusion allows tracking through visual occlusions of a leg, and that the sensors recovered the knee kinematics during this time (Figures \ref{fig:knee_angles} and \ref{fig:distributions}). 

Previous work has fused sensor and monocular video data; however, using stationary or calibrated cameras or calibrated IMUs. Work using a moving camera \cite{ferrari_recovering_2018} obtains 12 degrees of angular error across the whole body. This is not quite a fair comparison to the results presented here as their method evaluates the mean orientation error of SMPL body segments, which, as they overparameterize pose space, are likely to not be consistent between modalities. Our method for evaluating the knee angle does not have this overparameterization problem so it is not surprising that our method finds closer agreement. A strength of our method is it allows IMUs to be placed arbitrarily on arbitrary body segments. Recent work \cite{van_wouwe_diffusion_2023} allows for an arbitrary IMU placement as well and averages 13 degrees of error. OpenCap uses two calibrated stationary cameras \cite{uhlrich_opencap_2023} and finds joint orientation errors on par with our approach (4.1 degrees). OpenSense \cite{al_borno_opensense_2022} uses an explicit calibration step and finds MAE on par with our approach (3-6 degrees) and lower correlation coefficients (0.60-0.87). Another method, \cite{shin_markerless_2023} uses three multi-layer perceptrons to fuse IMU accelerometer, gyroscope and stationary video for full-body kinematics. While this method achieves results near 4 degrees of angular error, it still requires IMUs to be loosely calibrated to the body prior the to recording and does not utilize a moving camera. In summary, we find that our method combines the perks of all these methods: monocular, moving camera with arbitrarily placed, uncalibrated IMUs to find accurate, biomechanically grounded kinematics.

While we are excited about the ease of use of this method, it is not without limitations. First, the inclusion of a loss on the camera gyroscope may make this method less accessible, however, most smartphones do have a gyroscope and vision approaches can substitute for raw camera gyroscope values \cite{shin_wham_2023}. We acknowledge that our method only tracks orientation and translation relative to the smartphone camera. In cases where the experimenter remained stationary (Figure \ref{fig:example}), we saw a mean root translation error of 6 cm. We anticipate future work using the gait kinematics to inform the trajectory through the world can provide complementary information to track the person in world coordinates even in camera-translating conditions \cite{shin_wham_2023}.

Future work should apply this method to a larger clinical population. In particular, most lower limb prosthesis users had their device covered with their pant leg, which likely artificially increased the accuracy in that population. This work should also be validated on other joint kinematics as well, including the arm. Another future direction is fitting multiple trajectories simultaneously and sharing parameters like body scale and sensor orientation across the trajectories. We use this bilevel optimization approach \cite{werling_addbiomechanics_2023} in our
end-to-end optimization from the MMMC data, and find it is both accurate and reduces the time to fit multiple trajectories.

\section{CONCLUSIONS}
Sensor-body and camera calibration are time-consuming steps that limit the use of biomechanics and IMUs in routine clinical and at-home settings. We present a method to fuse handheld smartphone video and uncalibrated IMUs that produces accurate kinematics while retaining the high-resolution information in IMU readings. We also find that using an implicit function to reconstruct biomechanics from monocular video does remarkably well using modern 3D keypoint detectors and differentiable biomechanical models.

\addtolength{\textheight}{-12cm}   




\section*{ACKNOWLEDGMENT}

\printbibliography
\end{document}